\documentclass[11pt,a4paper]{article}
\usepackage[hyperref]{acl2019}
\usepackage{times}
\usepackage{latexsym}

\usepackage{url}
\usepackage{multirow}
\usepackage{graphicx}
\usepackage{amsmath}
\usepackage{verbatim}
\usepackage{pgfplots}
\pgfplotsset{compat=1.14}

\usepackage{color}
\usepackage{booktabs} 
\usepackage{subcaption}
\aclfinalcopy 
\def\aclpaperid{1224} 

\newcommand{\acldata}[0]{\emph{UPKConvArg}}
\newcommand{\aclrankdata}[0]{\emph{UPKConvArgRank}}
\newcommand{\ourdata}[0]{\emph{IBM-EviConv}}
\newcommand{\ourmethod}[0]{EviConvNet}

\definecolor{bblue}{HTML}{4F81BD}
\definecolor{rred}{HTML}{C0504D}

\title{Are You Convinced? Choosing the More Convincing Evidence with a Siamese Network}

\author{Martin Gleize$^{*}$, \enspace Eyal Shnarch\Thanks{ First two authors contributed equally.}, \enspace Leshem Choshen, \enspace Lena Dankin,  \\ 
 \enspace {\bf Guy Moshkowich,} \enspace {\bf Ranit Aharonov,} \enspace {\bf Noam Slonim} \\
 IBM Research \\
 martin.gleize@ie.ibm.com, guy.moshkowich@ibm.com, \\ \{eyals, leshem.choshen, lenad, ranita, noams\}@il.ibm.com}

\date{}

\begin{document}
\maketitle
\begin{abstract}
  
Machines capable of responding and interacting with humans in helpful ways have become ubiquitous. We now expect them to discuss with us the more delicate questions in our world, and they should do so armed with effective arguments. But what makes an argument more persuasive? What will convince you?

In this paper, we present a new data set, \ourdata{}, of pairs of evidence labeled for convincingness, designed to be more challenging than existing alternatives. We also propose a Siamese neural network architecture shown to outperform several baselines on both a prior convincingness data set and our own. Finally, we provide insights into our experimental results and the various kinds of argumentative value our method is capable of detecting.

\end{abstract}

\section{Introduction} \label{sec:intro}

The most interesting questions in life do not have a simple factual answer. Rather, they have pros and cons associated with them. When opposing sides debate such questions, each side aims to present the most convincing arguments for its point of view, typically by raising various claims and supporting them with relevant pieces of evidence. Ideally, the arguments by both sides are then carefully compared, as part of the decision process. 

Automatic methods for this process of argumentation and debating are developed within the field of Computational Argumentation. which focuses on methods for argument \emph{detection} \cite{Lippi2016,levy14,rinott15} and revealing argument \emph{relations} \cite{stab2014annotating,Stab2017ParsingAS}.

Recently, IBM introduced \emph{Project Debater}, the first AI system able to debate humans on complex topics. Project Debater participated in a live debate against a world champion debater, and was able to mine arguments and use them to compose a speech supporting its side of the debate. In addition, it was able to rebut its human competitor.\footnote{For more details and a video of the debate:\\ \url{https://www.research.ibm.com/artificial-intelligence/project-debater/live/}} The technology that underlies such a system is intended to enhance decision making. 

In this work we target the task of assessing argument \emph{convincingness}, and more specifically, we focus on evidence convincingness -- given texts representing evidence for a given debatable topic, identify the more convincing ones. 

Theoretical works have analyzed the factors that make an argument more convincing (e.g., \citealp{boudry2015fake}). This work is an empirical one in the line of \cite{persing2017can,Tan2016WinningAI}. To the best of our knowledge, this is the first work on evidence convincingness.

Most similar to our work is that of \citet{Habernal-ACL-16} who are the first to directly compare pairs of arguments (previous works compared documents). They released \acldata{}, the first data set of convincingness, containing argument pairs with a label indicating which one is preferred over the other. 

In this work we release \ourdata{}, a data set of evidence pairs which offers a more focused view of the argument convincingness task. As a source of evidence sentences we use the evidence data set released by \citet{BlendNet}, which contains more than 2,000 evidence sentences over 118 topics. We then sampled more than 8,000 pairs of evidence and sent them for convincingness labeling.

\paragraph{Why is the new data set useful?}
Our new data set differs from \acldata{} in a few important aspects. While in \acldata{} the pairs consist of two types or arguments, claims and evidence, \ourdata{} pairs are composed solely of evidence. In a follow-up work on \acldata{}, \citet{Habernal-EMNLP-16} showed that the most frequent reason by far to prefer one argument over another is that it is more informative. Usually, an evidence is longer and provides more details and information than a concise claim. Therefore, in a data set which includes both evidence and claims the identification of the more convincing argument may be based not only on argument convincingness, but also on identifying argument type, or even on a shallow feature such as argument length. Indeed, we show a very high performance of the baseline by length over \acldata{} in \S\ref{ssec:exp1}. On the other hand, a data set that includes only evidence poses a more challenging task. In addition, we directly controlled for argument length by building pairs of roughly the same length. 

A second important distinction between the data sets is writing level. The arguments for \acldata{} were extracted from two Web debate portals, on which people post free text and in which writing level widely varies (for instance, some posts include ungrammatical texts which require a pre-processing step). Our arguments were retrieved from Wikipedia, a heavily edited corpus which makes them on par in terms of writing level.

Overall, the contribution of this new data set is that it emphasizes pairs homogeneity -- in terms of argument type, length, and writing level. We believe that \ourdata{} offers learning algorithms a better chance to reveal real convincingness signals, beyond the more trivial ones.

Finally, \acldata{} pairs are of the same stance towards the topic, (either both supporting it or both contesting it), and therefore it is aligned with the task of choosing the most convincing arguments \emph{of a given side of the debate}. In contrast, our data set contains both same stance pairs, as well as cross stance pairs (i.e., one is supporting and the other is contesting the topic). Thus it is aligned with the above mentioned task, but in addition, with the task of choosing \emph{which side of the debate} was more convincing \cite{Potash2017TowardsDA}.

In addition to the release of a new data set, a second contribution of this work is the suggestion of a Siamese Network architecture for the argument convincingness task. We evaluate our method on both \acldata{} and \ourdata{} data sets, and show that it outperforms the methods suggested by \citet{Habernal-ACL-16} and \citet{Simpson-TACL-18} on both sets.

With the advancement in argument detection, the research community can now pay more attention to the challenging task of identifying the more convincing arguments. This work continues the line started by \citet{Habernal-ACL-16} by suggesting a focused framing of the task, providing a new data set for it, and presenting a neural network which surpasses state of the art performance.

\section{Background} \label{sec:bg}
\paragraph{Convincingness.}
Convincingness (or persuasiveness) arouses great interest in various fields such as essay scoring \cite{Ghosh2016CoarsegrainedAF}, persuasive technologies \cite{fogg1998persuasive,Fogg2002PersuasiveTU,Fogg2009ABM}, and social networks, where it is deemed a hard problem \cite{Hidey2018PersuasiveID}. 
Naturally, it is also relevant for social sciences, for example in public narrative \cite{Green2000TheRO}, internet discussions \cite{Tan2016WinningAI}, and in argumentative process of thought \cite{Burnstein2003PersuasiveAA}. 

In theoretical argumentation studies, the importance of quality \cite{wachsmuth2017argumentation} and convincingness was emphasized \cite{o2012conviction, van2014handbook, CKYCR19}, but assessment is still a challenge despite years of study \cite{weltzer2009assessing,rosenfeld2016providing}.

Traditionally, assessment of arguments convincingness, if addressed at all, relied on relevance, acceptability or sufficiency of arguments \cite{Habernal2015ExploitingDP,johnson2006logical}, or on general fallacies \cite{hamblin1971fallacies,tindale2007fallacies}. Recently, some works studied convincingness of full texts, assessing the role of prior beliefs \cite{Durmus2018EXPLORINGTR} and structure \cite{Wachsmuth2016UsingAM}.

\paragraph{Argument convincingness data set.}
Closer to our work, recent studies aim to assess the convincingness of a single argument, rather than that of a full text. The first data set for this task was published by \citet{Habernal-ACL-16}. Their data set, \acldata{}, consists of approximately 1,000 web mined arguments across 16 different topics, each split into two sets by stance (support or contest the topic). In each such split, all argument pairs are annotated by crowd workers for the preference of one argument over the other. In addition, the workers provided reasons for their choice in the form of free text. 

From the labeling over pairs, the authors proposed a method, based on PageRank \cite{page1999pagerank}, to derive a second data set, \aclrankdata{}, which approximates convincingness of individual arguments rather than in a comparative manner within an argument pair.  

A continuation work, on that data set, looks into the textual reasons provided by the annotators, classifies them and proposes prediction tasks on that classification \cite{Habernal-EMNLP-16}. 

\paragraph{Empirical methods for argument convincingness.}
To identify the more convincing arguments in a set we need to rank them. Learning to Rank is the machine learning field which aims to learn rankings rather than classification or regression \cite{McFee2010MetricLT,Burges2010FromRT}. Learning to rank can be formalized in various ways \cite{Cao2007LearningTR}; in a \textit{pointwise} approach, the input is single elements and for each the output is a score. To rank a list in this approach, one simply orders the elements by their scores. 

In a \textit{pairwise} approach, the input is pairs of elements and for each pair the output is the preference between its two elements. To rank a list in this approach, one must compare all pair combinations, assuming transitivity holds, otherwise some approximation is needed (such as the one made to produce \aclrankdata{}).

\citet{Habernal-ACL-16} suggest empirical methods for the task of choosing the more convincing argument. Relying on 32,000 linguistic features and word embedding, they proposed two methods, based on SVM and BiLSTM. When trained over argument pairs, these methods can provide pairwise inference only. They cannot predict a convincingness score for a single argument.

To overcome this disadvantage, \citet{Simpson-TACL-18} propose a pointwise algorithm based on Gaussian Process Preference Learning, \emph{GPPL}, \cite{Chu2005PreferenceLW} which is able to output a convincingness score per argument, while being trained on the pairs of arguments from \acldata{}. They use the same huge set of linguistic features and word embedding. They emphasize the importance of the pointwise approach, allowing for a more scalable and efficient inference. They also note that the benefits in avoiding neural networks lie in the superiority of graphical models for small training sizes like in \acldata{}. 

The Siamese network we propose next has all of those advantages and more. Being a neural network, it has the advantage of being more efficient in inference, and it can be updated with the frequent advances of this research field. In addition, the pre-processing step which generates the huge set of linguistic features, used by \citet{Habernal-ACL-16} and \citet{Simpson-TACL-18}, takes a lot of time and is not suitable for many languages. In contrast, our network does not depend on task specific features, and still achieves state of the art results on the task of argument convincingness classification and ranking.
\section{Siamese Network}\label{sec:mdl}
For the task of learning pointwise evidence convincingness scores from a data set of evidence pairs, we bring ideas from the field of learning to rank. Specifically, in our model, we take inspiration from  the training procedure provided by RankNet \cite{burges2005learning} of a Siamese network. Such a network consists of two legs of identical networks, which share all their parameters and are connected at the top with a softmax.

Unlike RankNet, we propose a network whose output is a probability. This is a desirable property as it is comparable with the output of other networks, and is understandable by humans. In initial experiments, on held-out data, the performance of our network was comparable to that of RankNet.

Each leg in our Siamese network is a neural network which is a function of an input argument $A$ and has two outputs $[C_A,D_A]$. $C_A$ represents how convincing $A$ is, and $D_A$ is a dummy output (which can be a constant). 

In training, given a pair of arguments, $A_i$ and $A_j$, we apply $\textrm{\em{softmax}}[C_{A_i},C_{A_j}]$, softmax over the convincingness output of each leg (ignoring the dummy output). The result is compared to the label of the pair using the cross entropy classification loss. Intuitively, this maximizes the probability of correctly identifying the more convincing argument, which pushes the margin between the two outputs to differ.

In inference, given a single argument, $A_k$, rather than a pair, the advantage of the Siamese network comes into play. To predict the convincingness score for $A_k$ we feed it into one of the legs. Then, we apply softmax, this time over the convincingness output and the untrained dummy output, $\textrm{\em{softmax}}[C_{A_k}, D_{A_k}]$. The higher the probability we get from this softmax, the more convincing $A_k$ is considered by the network.

\paragraph{Implementation of a leg.} Each leg in our Siamese network is a BiLSTM with attention as described in \citet{BlendNet}. We feed non-trainable word2vec embeddings \cite{mikolov2013distributed} to a BiLSTM of width 128, followed by 100 attention heads and a fully connected layer with two outputs. Training was done using Adam optimizer \cite{kingma2015Adam} with learning rate 0.001, applying gradient clipping above norm of 1 and dropout rate of 0.15. The system was trained for 10 epochs.
 \section{The \ourdata{} data set} \label{sec:data}
Following the motivation, presented in the introduction, for a more focused framing of the argument convincingness task, we release a new data set, \ourdata{}\footnote{
Available on the IBM Project Debater datasets webpage: \url{http://www.research.ibm.com/haifa/dept/vst/debating_data.shtml}}.

This data set is composed of 1,884 unique evidence sentences, extracted from Wikipedia, derived from the data set released by \citet{BlendNet}. These evidence spread over almost 70 different debatable topics. 

From the evidence set of each topic we sampled pairs, therefore within a pair, both evidence sentences refer to the same topic, arguing either for the topic (PRO) or contesting it (CON). In total we annotated more than 8,000 pairs, and after a cleaning step (detailed in \S \ref{ssec:data_clean}) we were left with 5,697 pairs that are split into train and test sets (4,319 and 1,378 pairs correspondingly). We kept the same split of \citet{BlendNet} in which no topic appears both in train and in test.  

The label of each pair indicates which evidence is more convincing out of the two. In addition, we provide the stance of each evidence towards the topic. 

Following is an example of a pair from our data set in which the first evidence was chosen to be more convincing: 
\begin{description}
\item[Topic:] We should legalize same sex marriage. 

\item [Evidence \#1:] The California Supreme Court overturned California's ban on gay marriages on May 15, stating that depriving gays and lesbians of the same rights as other citizens is unconstitutional. (PRO)

\item [Evidence \#2:] In his 2002 Senate campaign, Coleman pledged support for an amendment to the United States Constitution that would ban any state from legalizing same sex marriage. (CON)
\end{description}

Using Wikipedia as the source for evidence yields a data set that is rather homogeneous in vocabulary, grammar and style, as Wikipedia is heavily edited. In addition, as motivated in \S \ref{sec:intro}, we constructed pairs of evidences with roughly the same length, allowing for a length difference of up to $30\%$ of the shorter evidence. The evidence in each pair can have either the same stance or the opposite stance towards the topic. Overall, we annotated 3,075 pairs of the same stance towards the topic and 2,622 cross stance pairs. 

Each pair in \ourdata{} was annotated by 10 crowd labelers.\footnote{we used the Figure-Eight platform: \url{https://www.figure-eight.com/}} Out of all the pairs a labeler annotated, $20\%$ were \emph{hidden test questions} used to verify annotations quality (see \S\ref{ssec:data_clean}). 

The labelers were provided with the following guideline, and were asked to be decisive: \\ 
\textit{In a conversation about the topic, where you can only give a single evidence out of the following two - which one would you rather use? Which one is more convincing?} 

We consider an evidence to be more convincing than its counterpart if it was chosen by at least $60\%$ of labelers. Pairs in which one evidence was preferred by more than half of the labelers but less than this threshold were considered as indecisive and were removed from the data set. 

After data set cleaning, described next, the most frequent label in the data set (train and test sets together) covers only $53\%$ of the pairs. Hence, it is safe to say the data set is balanced and there is no strong bias towards a certain sentence length. 


\subsection{Data set quality} \label{ssec:data_clean}
We took several measures to ensure the quality of \ourdata{}. First, we selectively picked crowd labelers based on their performances and credibility on previous tasks of our team. In total, 92 labelers of this group took part in the annotations. 

We initially performed a pilot annotation task to evaluate the quality of the crowd work by comparing their annotations to those of in-house expert labelers. The pilot contained $105$ pairs that were labeled by both groups. We filtered out $21$ pairs whose labeling was indecisive by either group (see previous section) and found out that for $84\%$ of the remaining pairs the two independent groups agreed on the label. This encouraging result indicated that the crowd labelers are suitable for this annotation task.

We further filtered specific crowd labelers whose work did not adhere the following requirements: (i) annotating a minimal number of 20 pairs, (ii) obtaining a minimal average Cohen's Kappa \cite{Cohen1960} of $0.1$ calculated over enough shared content with other labelers (i.e., sharing at least 20 pairs with at least $10$ labelers). 

Another filter mechanism is based on hidden test questions. These test questions were constructed automatically by pairing a confirmed evidence with a rejected evidence candidate from the data set of \citet{BlendNet}. Naturally, the confirmed evidence is labeled as the more convincing one, since the other sentence is not even a proper evidence. These questions were randomly placed among the true pairs for annotation and the labelers could not know which question was a test one (therefore are called hidden). A labeler whose precision over these test questions was below $0.55$ was filtered out. The average precision of the remaining labelers over the hidden test questions is $0.73$.

In total, we filter 23 labelers by these criteria, and removed all of their annotations. Following this process, pairs that were left with less than 7 annotations by valid labelers were removed from the data set to maintain a high standard of majority.

We use Cohen's Kappa to calculate the average pairwise agreement of the labelers, yielding the score of $0.33$. 
We note that the average pairwise Cohen's Kappa score of our expert labelers was $0.38$ on this task, indicating the difficulty of this task to humans. This agreement level is a typical value in such challenging labeling tasks (e.g., \citealp{aharoni14}). We consider it an upper bound and therefore see the $0.33$ Kappa score of our crowd labelers as an acceptable agreement.

Finally, a desired characteristic of such a data set is that transitivity among evidence pairs will hold \cite{Habernal-ACL-16}. This is our last quality test of \ourdata{}. We extract all 1,899 triplets of evidence for which all pairs were annotated. Then, we calculate the percentage of such triplets in which transitivity holds, i.e. one evidence is consistently considered the most convincing, one the least convincing and the third is in the middle. The results were surprisingly high, $99\%$ of the triplets comply with the transitivity expectation.
\section{Experiments} \label{sec:exp}

From this point on, we will refer to the method presented in \S\ref{sec:mdl} as \emph{\ourmethod{}}.

\subsection{Experiments over \emph{UKPConvArgStrict} and \emph{UKPConvArgRank}} \label{ssec:exp1}

We first experiment on the argument convincingness data set released by \citet{Habernal-ACL-16}. It is split into two tasks: pair classification on \textit{UKPConvArgStrict}, and ranking on \textit{UKPConvArgRank}. On UKPConvArgStrict, all systems were evaluated in cross-topic validation over 32 topics (16 actual topics, with 2 stances each) and their average accuracy across folds is reported in Table \ref{tab:exp1-classification}.

\begin{table}[ht]
\begin{center}
\begin{tabular}{lc}
  System & Accuracy\\
  \midrule
    Most frequent label & 0.50 \\
	BiLSTM & 0.76 \\
	Argument length & 0.77 \\
	SVM	& 0.78 \\
	GPPL opt. & 0.80  \\
	GPC & \textbf{0.81} \\
	\ourmethod{} & \textbf{0.81} \\
\end{tabular}
\end{center}
\caption{Accuracy on \emph{UKPConvArgStrict}. Our model (\ourmethod{}) is comparable to the best baseline.}
\label{tab:exp1-classification}
\end{table}

\textit{BiLSTM} and \textit{SVM} are the methods presented in the original paper \cite{Habernal-ACL-16}. \textit{GPPL opt.} and \textit{GPC} are Gaussian process methods later demonstrated by \citet{Simpson-TACL-18}. Our own \ourmethod{} performs similarly to the best previously known systems on this task.

Most of these systems were also evaluated on UKPConvArgRank, where each single argument is assigned a score (yielding a ranking). The original work reported Pearson's $r$ and Spearman's $\rho$ on the combined ranking of all arguments from all 32 topics. Subsequent work \cite{Simpson-TACL-18} reported the average of these measures across topics. We report their results and ours in this setting in Table \ref{tab:exp1-ranking}. Again \ourmethod{} provides at least equal performance to the best previously known method, \textit{GPPL opt}, and a statistically significant increase in Pearson's $r$ ($p \ll 0.01$ using one-sample two-tailed t-test).

\begin{table}[ht]
\begin{center}
 \resizebox{\linewidth}{!}{
\begin{tabular}{lcc}
  System & Pearson's $r$ & Spearman's $\rho$ \\
  \midrule
 	Argument length & 0.33 & 0.62 \\
 	BiLSTM & 0.36 & 0.43 \\
	SVM	& 0.37 & 0.48 \\
	GPPL opt. & 0.44 & \textbf{0.67} \\
	\ourmethod{} & \textbf{0.47} & \textbf{0.67} \\
\end{tabular}
}
\end{center}
\caption{Correlation measures on UKPConvArgRank. Our model (\ourmethod{}) is comparable to the best baseline.}
\label{tab:exp1-ranking}
\end{table}

We also tested a simple baseline assigning the argument's character length as its score (\textit{Argument length}). In pair classification, the baseline prefers the longer argument. We noted performances comparable to the original method of \citet{Habernal-ACL-16}. This result is in line with what the authors reported in a further study of the reasons given by annotators for preferring one argument over the other \cite{Habernal-EMNLP-16}: the most common reason provided is by far "more details, information, facts, examples / more reasons / more specific".

\subsection{Experiments over \ourdata{}} \label{ssec:exp2}

We report in Table \ref{tab:exp2} the accuracy of various baselines and our own method, on the full \ourdata{} data set. 

The simplest baseline is preferring the longest sentence, as before, but on this data set it has nearly the same accuracy as just picking the first candidate every time (\textit{most frequent label}). 

The \textit{Detection model} assigns a score to each individual sentence, and we choose the sentence with the highest score. To produce this score, a single leg of the network presented in Section \ref{sec:mdl} is used, with a softmaxed 2-dimensional output, trained using cross entropy classification loss over evidence candidates in the base data set from \citet{BlendNet}.\footnote{Detection scores are provided with the data set we release for ease of reproducibility.} 

We also run the GP-based methods proposed by \citet{Simpson-TACL-18}\footnote{Using their code from \url{https://github.com/ukplab/tacl2018-preference-convincing}.}. The increase these methods bring over the detection model is statistically significant ($p \ll 0.01$ using Wilcoxon signed-rank test). \ourmethod{}, the Siamese network described in \S\ref{sec:mdl}, significantly outperforms all systems ($p \ll 0.01$). We note that the gains from better methods are far greater here than in \S\ref{ssec:exp1}: GPPL improves over the sentence length baseline by 26\% and our method improves over GPPL by 9\% on \ourdata{}, compared to improvements of only 5\% and 1\% on \acldata{}.\footnote{Percentages are relative to the accuracy of the system or baseline referred to. This is to allow a more meaningful comparison of behavior on the two datasets.}

\begin{table}[ht]
\begin{center}
\begin{tabular}{lc }
  System & Accuracy\\
  \midrule
 	Evidence length & 0.53 \\
	Most frequent label	& 0.54 \\
	Detection model 	& 0.59 \\
	GPPL & 0.67  \\
	GPPL opt. & 0.67  \\
	GPC & 0.67 \\
	\ourmethod{} & \textbf{0.73} \\
\end{tabular}
\end{center}
\caption{Accuracy on \ourdata{}. our model (EviConvNet) outperforms prior art and our additional baselines.}
\label{tab:exp2}
\end{table}
\section{Analysis} \label{sec:analysis}
In this section we present an analysis of several interesting aspects of our new data set and method.

\subsection{Performance across preference reasons}
\citet{Habernal-EMNLP-16} analyze and categorize the reasons provided by the labelers of \acldata{} to justify their choice on each pair (see Table \ref{tab:reasons} for examples of reasons). The most common reason is that an argument is more informative (code C8-1 in the table). As valid and pervasive as this factor is in real arguments, it also makes the argument length a high-performance baseline which is hard to beat (as seen in \S\ref{ssec:exp1}).

One of the motivations for our work was to create another data set where the amount of textual content would not be a factor in the choice of labelers, possibly constraining the preference task to the more subtle aspects of ``convincingness''. 

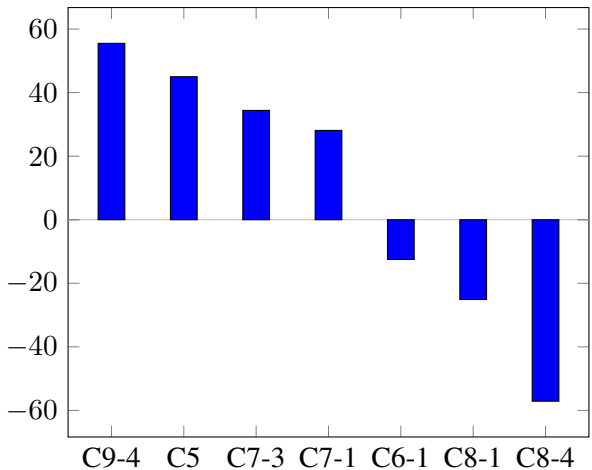
\begin{figure}[htb]
\begin{tikzpicture}
\begin{axis}[
    symbolic x coords={C9-4,C5,C7-3,C7-1,C6-1,C8-1,C8-4},
    extra y ticks       = 0,
    extra y tick labels = ,
    extra y tick style  = { grid = major },
    xtick=data]
    \addplot[ybar,fill=blue] coordinates {
        (C9-4,55.5)
        (C5,45)
        (C7-3,34.4)
        (C7-1,28.1)
        (C6-1,-12.5)
        (C8-1,-25.1)
        (C8-4,-57.1)
    };
\end{axis}
\end{tikzpicture}
\caption{Relative decrease in error rate (\%) between argument length baseline and \ourmethod{} (reason codes defined in Table \ref{tab:reasons}).}
\label{fig:reasonrelative}
\end{figure}

We compute the error rate ($1$ - accuracy) of the length baseline and \ourmethod{} on pairs clustered by their \textit{reason units} as defined by \citet{Habernal-EMNLP-16}. For clarity of the analysis, the pairs were restricted to those where a single reason was given. We selected the four reasons presenting the highest relative decreases in error rate and the three single reasons where the baseline outdoes our method. Figure \ref{fig:reasonrelative} shows the relative decrease in error rate between argument length baseline and \ourmethod{}, with the reason codes defined in Table \ref{tab:reasons}.

We note that unsurprisingly, our neural network model does better than the length baseline at capturing what an argument should be like in term of presentation, relevance and quality of content.

\begin{table}[htb]
    \centering
     \resizebox{\linewidth}{!}{
    \begin{tabular}{ll}
        Code & Reason \\
        \midrule
        C9-4 & Well thought out / higher complexity \\
        C5 & Language / presentation of argument \\
        C7-3 & Off-topic / doesn't address the issue \\
        C7-1 & Not an argument / is only opinion / rant \\
        \midrule
        C6-1 & Not enough support / not credible \\
        C8-1 & More details, information, examples \\
        C8-4 & Balanced, objective, several viewpoints \\
    \end{tabular}
    }
    \caption{Reasons for choosing a better/worse argument taken from \citet{Habernal-EMNLP-16}.}
    \label{tab:reasons}
\end{table}

More interesting are the reasons where the neural network does not perform as well. Two of those are about the sheer amount of supporting information, which would indeed be more directly captured by the length baseline. The reason where \ourmethod{} has a $57\%$ greater error rate is about presenting a balanced, objective argument which tackles different viewpoints. These pairs only make up 3\% of the data set, so it is possible the network needed to see more such pairs in training to perform well on them. 

\subsection{What makes a convincing evidence?}

\paragraph{We asked the experts.} We asked our in-house expert labelers to supply factors for commonly deciding their preference (their answers are released with the data set). Reliability of the source was an important factor, including titles and names, level of expertise, type of evidence (study, expert, opinion, example, precedent), whether the source has an interest in the discussed matter, and where it came from geographically. 

Also important were content issues, whether information was complete, specific, significant, rhetorically strong, the amount of supporting evidence or details reported and the relevance of the evidence to the present, and better yet to the future or in general. Some technical issues were also reported, such as missing information or an opinion rather than a fact.

Additionally we inquired about the cases that were difficult to compare. These tended to be either cases where both
pieces of evidence were not convincing (for the
reasons above), or where it was hard to ascertain -- for
a certain factor (e.g. reliability or significance) --
which argument prevailed.

\paragraph{We asked the network.}
\begin{figure*}[t]
	\begin{subfigure}[b]{0.5\textwidth}
	    \includegraphics[width=\columnwidth]{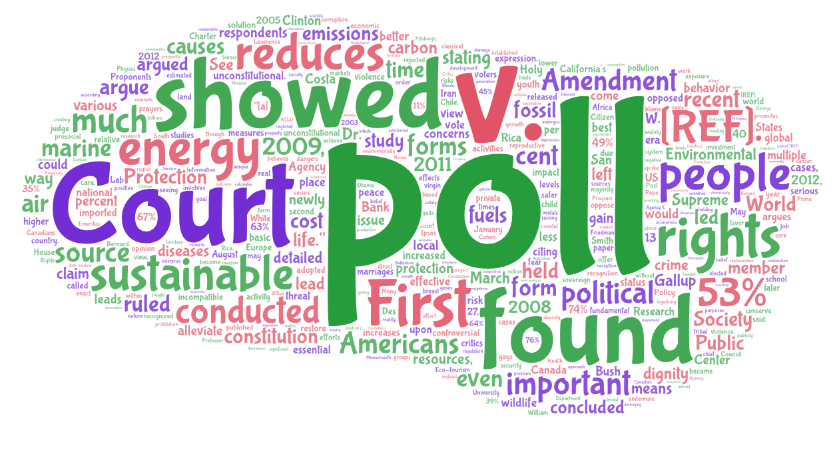}
	    \caption{true convincing dist - true non-convincing dist}
	    \label{subfig:TP-TN}
    \end{subfigure} 
    \begin{subfigure}[b]{0.5\textwidth}
	    \includegraphics[width=\columnwidth]{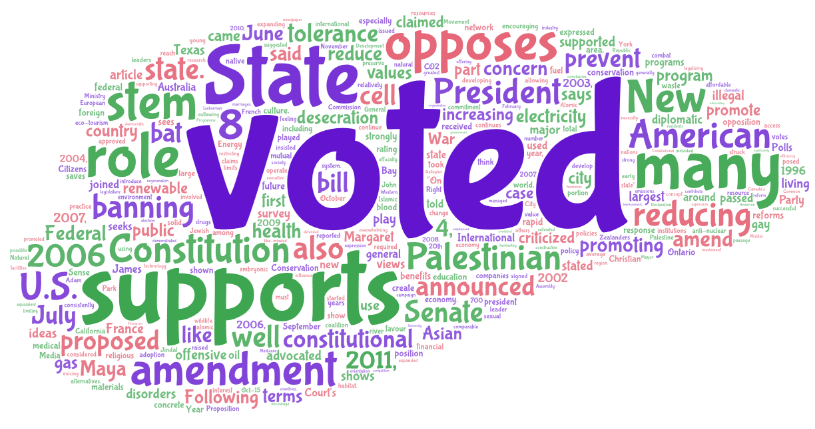}
	    \caption{true non-convincing dist - true convincing dist}
	    \label{subfig:TN-TP}
    \end{subfigure}
\caption{Word clouds of correctly classified pairs. 
\label{fig:clouds}}
\vspace{-0.3cm}
\end{figure*}

To acquire a better understanding of what typical things differentiate convincing evidence from non convincing ones, we compared word distributions on pairs that our network successfully classified (Figure \ref{fig:clouds}). From the correctly classified pairs we construct two sets; one is composed of the more convincing evidence in each pair, \emph{true convincing}, and the other contains the less convincing ones, \emph{true non-convincing}.

For each set, we calculate the distribution of unigrams in its evidence sentences, ignoring stop words and unigrams which appear in the topic title. In Figure \ref{fig:clouds} we present the differences between the two distributions, thus, discarding words that are common in many evidence sentences regardless of the convincingness of their texts. 

On the left side of Figure \ref{fig:clouds} we see the words which are much more prominent in convincing arguments than they are in non-convincing ones. We find there words related to argumentation (\textit{argue}, \textit{claim}), or studies and polls (\textit{found}, \textit{conducted}, \textit{[REF]\footnote{A sign which indicates that this evidence was taken from a written source (it replaces the reference text to the source).}}). 
Other words mention authoritative figures (\textit{DR.}, \textit{Clinton}, \textit{ W. Bush}) or court orders (\textit{supreme}, \textit{v.}\footnote{V. is used as a versus abbreviation in court rulings.}).

On the other hand, when subtracting the true convincing distribution from the true non-convincing distribution (Figure \ref{subfig:TN-TP}) one gets 
opinion words (\textit{support}, \textit{opposes}, \textit{vote}), 
partial change (\textit{reduce}, \textit{amend}, \textit{part}), 
non-emphasized actions (\textit{said}, \textit{proposed}, \textit{concern}).

\subsection{Cross vs. same stance evidence pairs}
As described in \S\ref{sec:data} we build our data from pairs with the \emph{same-stance} towards the topic, as well as \emph{cross-stance} pairs, in which one argument supports the topic while the other opposes it. We created the cross-stance pairs since we had in mind the task of comparing arguments of different sides of a debate. Given this task, some questions naturally arise, such as
\begin{itemize}
    \item Is it a harder task to identify the more convincing argument when comparing arguments of opposite stances?
    \item Is it better to train on cross-stance pairs for this task?
\end{itemize}

With \ourdata{} we can empirically examine such questions.

To this end, we extracted three subsets from the training set (of 2,082 pairs each); one with same-stance pairs only, the second with cross-stance pairs only, and the third with mixed-stance. Similarly we extracted three subsets from the test set (each with 385 pairs). 

Given these data sets we are able to test what happens when we train and test our network on all combination of pairs of same/cross/mixed stance. 

\begin{table}[ht]
\begin{center}
\begin{tabular}{l l  c  c  c  }
 \multicolumn{2}{c}{} & \multicolumn{3}{c} {Test} \\
 & & same & cross & mixed \\
   \toprule
 \multirow{3}{*}{Train} & same & 0.72 & 0.71 & 0.71 \\
                        & cross & 0.72 & 0.69 & 0.71 \\
                        & mixed & 0.72 & 0.71 & 0.70 \\
 \bottomrule
\end{tabular}
\end{center}
\caption{Comparing accuracy when training and testing on each combination of same/cross/mixed stance.}
\label{tab:same_cross_stance}
\end{table}

Table~\ref{tab:same_cross_stance} depicts the results. To our surprise, training on cross-stance pairs does not improve performance on a test with cross-stance pairs in comparison to training on same or mixed stance pairs (middle column in the table). Same goes for the other subset. In addition, it appears that cross-stance pairs do not pose a more difficult task than same-stance pairs or mixed-stance, as the accuracy over them is not smaller than the accuracy on same-stance or mixed-stance pairs.

\subsection{The effect of length difference}
In previous sections, we discussed our choice to limit the difference in length between evidence of the same pairs. This decision was encouraged by the relatively good results of the argument length baseline on the \acldata{}. Still, one may wonder whether training on similar length pairs harms the performance over real life pairs in which length balance is not guaranteed. For that purpose we annotated $458$ pairs with a significant length difference (higher than $30\%$, complementary to the restriction in \ourdata{}). 

The accuracy of \ourmethod{} over this test set is $0.69$, which is lower than $0.73$, the accuracy over the balanced data set, reported in Table \ref{tab:exp2}, but still higher than all other baselines. This difference is small enough to conclude that our model, trained on a length balanced data set generalizes well enough to be able to classify pair of evidence of different lengths. 
\section{Discussion and future work}
In this work we proposed a focused view for the task of argument convincingness, constructed a new data set for it, and presented its advantages. We believe that it is useful to evaluate methods for identifying the more convincing argument on this more challenging data set. 

In addition, we suggest our version of a Siamese network for the task, which outperforms state of the art methods.     

A possibility that we did not expand on in this paper is to pre-train one leg of the network on an argument detection data set, like the one of \citet{BlendNet}. Argument detection concerns itself with the binary classification of a single text into argument and non-argument, and not the more subjective notion of convincingness. But we nonetheless observed in previous experiments significant improvements when initializing the Siamese network with weights learned on this task. We could not reproduce these improvements here, but our previous efforts relied on far fewer training pairs: an explanation could be that pre-training is most helpful when faced with a low amount of training data.

In the future we aim to test and adapt other improvements in the learning to rank field to our task, hoping for further improvement by those models \cite{Burges2010FromRT,severyn2015learning}. In addition, more careful design of the architecture details, which was not the focus of this work, will probably yield better results yet, e.g., contextualized word embeddings \cite{peters2018deep}, batch normalization \cite{ioffe2015Batch,cooijmans2017Recurrent}, deeper networks and other architecture practical heuristics.


\bibliography{0.main}
\bibliographystyle{acl_natbib}


\end{document}